\documentclass[11pt]{article}

\usepackage[preprint]{acl}

\usepackage{times}
\usepackage{latexsym}

\usepackage[T1]{fontenc}

\usepackage[utf8]{inputenc}

\usepackage{microtype}

\usepackage{inconsolata}

\usepackage{graphicx}

\usepackage{amsmath, amssymb}
\usepackage{multirow, makecell}
\usepackage[ruled,vlined,linesnumbered]{algorithm2e}
\usepackage{amsthm}
\newtheorem{proposition}{Proposition}
\newtheorem{remark}{Remark}

\usepackage{booktabs}

\title{Debate to Deceive: Poisoning LLM's Fact-Checking via Game-Theoretic Multi-Agent Debate}
\author{
  \textbf{Liting Huang\textsuperscript{1}},
  \textbf{Huizhi Liang\textsuperscript{1}},
  \textbf{Yi Lei\textsuperscript{2}},
  \textbf{Xingjun Ma\textsuperscript{3}},
  \textbf{Varun Ojha\textsuperscript{1}}
\\
\\
  \textsuperscript{1}Newcastle University, United Kingdom,
  \textsuperscript{2}Peking University, China,
  \textsuperscript{3}Fudan University, China
\\
}

\begin{document}
\maketitle
\begin{abstract}
Knowledge attacks on Large Language Models, such as manipulating fact-checking judgments, pose significant risks to AI reliability. Existing approaches—adversarial suffix/template injection and single-agent self-reflection—lack robustness across diverse defenses and are prone to confirmation bias and costly trial-and-error, respectively. We propose \textsl{Debate-to-Deceive (D2D)}, a multi-agent debate framework grounded in human debate structure and rhetorical style, enabling agents to generate adversarial passages that are semantically coherent, rhetorically effective, and resistant to early confirmation bias. D2D formalizes the adversarial process as an evolutionary game in which agents adaptively evolve their rhetorical strategies (Logos, Pathos, Ethos) through competition. To manage the intractably large LLM output space, we decouple agent behavior into a strategic layer, which selects a rhetorical action from a compact strategy space via evolutionary-inspired learning dynamics, and a generation layer, where the LLM produces a concrete passage conditioned on that strategy and the debate context. On four fact-checking benchmarks—FEVER, HealthVer, CLIMATE-FEVER, and SciFact—D2D substantially outperforms existing attacks and evades most defenses.
\end{abstract}

\section{Introduction}
\label{sec:intro_d2d}
Large Language Models (LLMs) are increasingly used for reasoning, knowledge retrieval, and decision support in professional, legal, and social domains~\cite{luo2022biogpt, zhou2025rulearena, liuautodan}. As a result, ensuring their factual reliability and robustness against adversarial manipulation has become critical. Knowledge attacks, especially those targeting fact-checking judgments, can mislead aligned LLMs into hallucinating, contradicting established facts, or bypassing safety guardrails~\cite{tu2025knowledge, wu2025admit, liuautodan, liu2024formalizing, shi2024optimization, rahman2025fine, woodualfocus, esugo2025shield, alon2023detecting}.

Existing adversarial attacks can be broadly grouped into three categories. Gradient-optimized suffix attacks use token-level optimization to induce harmful outputs~\cite{shi2024optimization, zou2023universal, guo2021gradient}. Manual or template-based jailbreak attacks rely on crafted prompts, paraphrases, or distracting instructions to override model alignment~\cite{liu2024formalizing, zhang2025practical, shen2024anything, wei2023jailbroken, kang2024exploiting, zou2025poisonedrag}. LLM-based iterative attacks refine adversarial passages through feedback or search; for example, AutoDAN uses a hierarchical genetic algorithm~\cite{liuautodan}, while ADMIT introduces memory-based self-reflection during adversarial generation~\cite{wu2025admit}.

However, existing attack methods still face key limitations in robustness, efficiency, and cost. First, gradient-based attacks produce adversarial suffixes composed of meaningless character sequences~\cite{zou2023universal, morrisetal2020reevaluating}, which are easily flagged by perplexity-based filters or known-answer defenses~\cite{jain2023baseline, wang2024defending, phute2024llm, liu2024formalizing, instruction_defense}. Second, manual or template-driven attacks rely on a narrow set of fixed prompts, limiting the diversity of adversarial passages and their robustness across defenses. Third, LLM-based iterative optimization attacks remain constrained: ADMIT~\cite{wu2025admit} leverages inference-time memory (e.g., summarizing past interactions) but, as a single-agent method, lacks adversarial correction and is prone to confirmation bias and costly trial-and-error. AutoDAN~\cite{liuautodan} improves over fixed templates via a hierarchical genetic algorithm, yet incurs a large GPU memory footprint and remains fundamentally template-based, leaving it vulnerable to known-answer detection. This raises a central question: \textit{Can semantically coherent, rhetorically effective adversarial passages be generated at low cost and high speed, while remaining robust across diverse defenses?}

To address this question, we draw on the structure of human debate competitions as a framework for efficient, adaptive adversarial generation. Human debate takes multiple forms~\cite{gray1926oregon, nsda2024worldschoolsfinal, ICDC}: the Oregon format structures one-on-one debate into three phases~\cite{gray1926oregon}, while the Singapore format structures four-person team debate into four phases—foundational reasoning (definitions and facts), associative reasoning (standard arguments and counterarguments), core reasoning (novel insights and impactful rebuttals), and conclusive reasoning (synthesis)~\cite{ICDC}. These structures are complemented by the classical rhetorical triad—Logos, Pathos, and Ethos—which governs how arguments are framed in terms of logic, emotion, and credibility. Inspired by this structure, we propose Debate to Deceive (D2D), an attack framework that formalizes adversarial generation as a structured debate grounded in an evolutionary game formulation and multi-agent reinforcement learning (MARL). D2D adapts the four-phase Singapore format to one-on-one debate, using multiple agents that dynamically compete to generate adversarial passages that are both semantically coherent and rhetorically effective—in contrast to prior methods relying on fixed suffixes or single-agent optimization. D2D is motivated by three design considerations: 1) effective deception requires dynamic rhetorical adaptation, as static strategies are easily detected; 2) single-agent self-correction is prone to confirmation bias and costly trial-and-error, whereas multi-agent debate improves correction quality and reduces search cost; and 3) debate is inherently an adaptive strategic process in which argument effectiveness improves through competition, motivating an evolutionary game formulation with MARL to model this dynamic. The main contributions of this work are threefold:
\begin{itemize}
    \item \textbf{Structured debate-based attack framework.} We introduce D2D, a multi-agent adversarial framework inspired by human debate competition, combining staged reasoning with rhetorical strategies such as Logos, Pathos, and Ethos to manipulate LLM fact-checking judgments.
    
    \item \textbf{Evolutionary game formulation with multi-agent reinforcement learning (MARL):} We cast adversarial generation as an evolutionary game in which agents co-evolve their rhetorical strategies. We use a hierarchical pipeline where agents first choose a rhetorical strategy, then generate text conditioned on that choice. 
    
    \item \textbf{Demonstration of strategic vulnerability across diverse defenses:} We show that diverse defense methods remain vulnerable to our D2D-based debate-generated adversarial samples as they can induce erroneous judgments into the target LLM.
\end{itemize}

\section{Related Work}
\label{sec:related_work_d2d}
\subsubsection*{Game theory in LLMs.} Recent LLM research has largely adopted (behavioral) game theory for capability assessment and mechanism-oriented improvement. Hua et al.~\cite{hua2024game} evaluate LLM rationality in complete- and incomplete-information games and propose game-theoretic workflows to improve equilibrium reasoning and robustness. Fan et al.~\cite{fan2024can} operationalize rationality as a core metric and systematically characterize LLM behavioral boundaries across canonical games. Akata et al.~\cite{akata2025playing} study finitely repeated $2\times2$ games played by LLMs against other LLMs and humans, identifying systematic cooperation/coordination patterns and showing that behavior can be modulated via prompting. Mao et al.~\cite{mao2025alympics} propose \textsl{ALYMPICS}, an empirical game-theory framework that uses LLM agents as experimental subjects to support controlled simulation studies (e.g., multi-round resource auctions). Sun et al.~\cite{sun2025game} provide a systematic survey of the bidirectional interplay between game theory and LLMs, covering benchmarks, algorithmic improvements, and societal-impact modeling. Overall, existing LLM-related studies mainly use game theory to evaluate, characterize, or scaffold LLM behavior under predefined game rules and payoff structures. In contrast, our work uses game-theoretic payoff as an internal adaptation mechanism for adversarial generation. Rather than inheriting the payoff structure from a canonical game, we explicitly design the strategic payoff and combine it with rewards for real-time generation quality and cost, enabling rhetorical strategies to evolve over repeated multi-agent interactions.

\subsubsection*{Multi-agent debate.} Multi-agent debate has emerged as an effective framework for improving LLM reasoning, factuality, and decision quality through structured interaction among multiple agents. Du et al.~\cite{du2023improving} show that debate-based interaction can improve factuality and reasoning in language models. Khan et al.~\cite{khan2024debating} further demonstrate that stronger and more persuasive debaters can lead to more truthful answers, highlighting the importance of persuasive dynamics in debate. Beyond accuracy-oriented reasoning, Liang et al.~\cite{liang2024encouraging} use multi-agent debate to encourage divergent thinking, while Yi et al.~\cite{yidebate} incorporate belief modeling and Bayesian Nash equilibrium into debate-based reasoning. Estornell and Liu~\cite{estornell2024multi} provide a broader framework for analyzing principles and interventions in multi-LLM debate. Despite these advances, existing work mainly studies debate as a mechanism for collaborative reasoning, truth-seeking, or diversity enhancement. By contrast, our work focuses on competitive multi-agent debate, where agents pursue conflicting objectives and adapt their rhetorical strategies over time. Rather than using debate solely to improve the correctness of answers, we formalize adversarial persuasion as a strategic interaction in which agents infer opponents' tendencies and update their strategies to maximize persuasive effectiveness.

\subsubsection*{Attacks against LLMs.}
Adversarial passages can be categorized into two types: those with meaningless prefixes/suffixes and those with semantically meaningful adversarial passages. \textit{Jailbreaking attacks} aim to bypass safety mechanisms and elicit harmful or prohibited responses from LLMs. These attacks can be divided into white-box and black-box approaches. White-box attacks require access to the model's internal parameters or gradients. Gradient-based optimization methods, such as GCG~\cite{zou2023universal}, compute gradients to systematically optimize adversarial prompts by accessing the model's output probability distributions (logits). Training-based approaches, such as shadow alignment~\cite{yang2024shadow} and weak-to-strong generalization~\cite {burnsweak}, exploit model training dynamics to generate jailbreak prompts. In contrast, black-box attacks operate without access to model internals, relying only on input-output interactions. LLM-based generation methods such as AutoDAN~\cite{liuautodan}, JudgeDeceiver~\cite{shi2024optimization}, PAIR~\cite{chao2025jailbreaking}, and PAP~\cite{zeng2024johnny} use LLMs to automatically generate adversarial inputs that bypass safety filters. Another line of work, knowledge-to-jailbreak~\cite{tu2025knowledge}, collects large-scale datasets of knowledge-jailbreaking pairs and fine-tunes an LLM as a specialized jailbreak generator to produce domain-specific adversarial prompts. \textit{Prompt injection attacks} manipulate the behavior of LLM by injecting malicious instructions into user prompts. Direct prompt injection approaches, such as Open-Prompt-Injection~\cite{liu2024formalizing}, design carefully crafted prompts that override the model's original instructions or safety constraints. Memory-based injection methods with proxy passages, such as ADMIT~\cite{wu2025admit}, leverage iterative self-refinement through memorized proxy content to achieve adversarial objectives. Overall, existing attack methods mainly optimize prompts or generators to produce effective adversarial inputs. By contrast, our work focuses on how multiple agents strategically adapt rhetorical strategies during adversarial generation. Rather than treating attack construction as a single-prompt optimization problem, we model it as a dynamic 
multi-agent process in which persuasion strategies evolve based on strategic feedback and generation quality.

\section{Methodology}
\label{sec:method_d2d}
\subsection{Notations}
\begin{figure}[ht]
  \centering
  \includegraphics[width=1.0\linewidth]{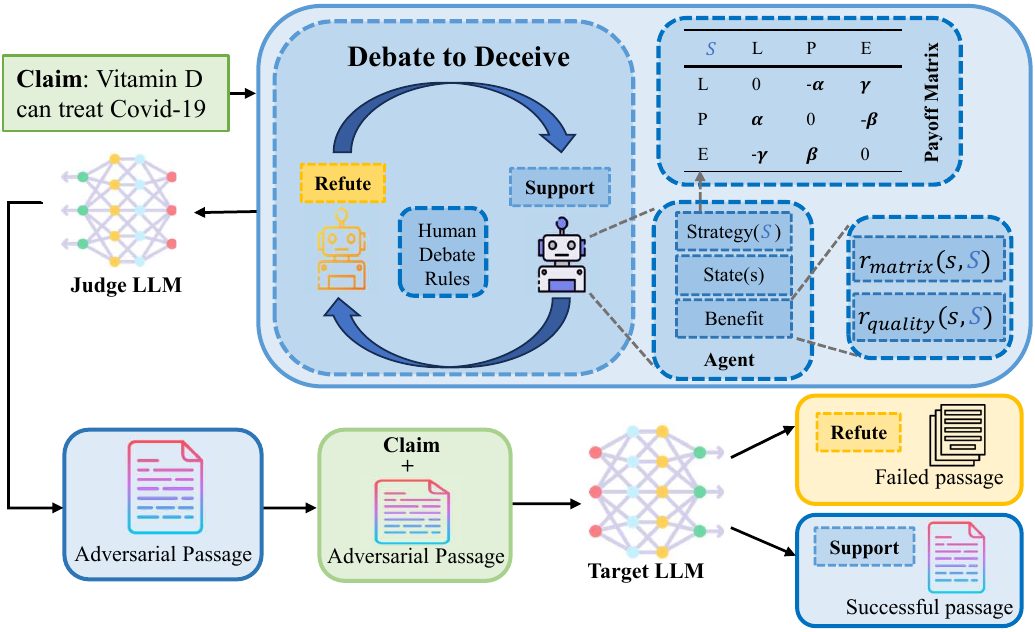}
  \caption{D2D framework: Schematic diagram of a debating adversarial attack framework targeting an LLM for fact-checking. An attacker orchestrates two AI agents to engage in a structured human debate, iteratively crafting adversarial queries that are imperceptible to humans but effective in misleading the target LLM. This process induces incorrect responses. For example, given the claim ``Vitamin D can treat COVID-19,'' D2D generates adversarial passages that steer the target LLM toward a malicious fact-checking outcome.}
  \label{fig:D2D_Framework}
\end{figure}
D2D operates in a black-box setting: the attacker has no access to model parameters and injects adversarial text directly into the evidence field while observing output verdicts. Our threat model has the following: 1) attacker capability: black-box, query-based; 2) attacker constraint: fluent natural language, no gradient access; 3) attack surface: direct evidence injection. A successful attack induces the target LLM to generate poisoned outputs. Thus, given a claim $C$ to be fact-checked, the attacker's objective is to: 1) flip the original verdict (e.g., Support $\rightarrow$ Refute), and 2) generate an adversarial passage consistent with the flipped outcome. Such adversarial passages can mislead the target LLM into producing an incorrect fact-checking response.

In this context, let $\mathcal{D} = (C, L, E)$ denote fact data, where $C = \{c_1,c_2, \ldots, c_n\}$ is the set of $n$ claims, $L = \{l_1,l_2, \ldots,l_n \}$ is the true label of the claims, and $E$ is the evidence of the claim in the correct label side. We define the target label as $L^{'}$, which is the opposite of label $L$. We focus on agent debate, in which one agent argues for the original label of a claim and the other for the flipped (target) label. These two agents follow the structure of a human debate competition game to propose an opinion, including foundational, associative, core, and conclusive reasoning. In this setting, $A = \{A_L, A_{L'}\}$ is a pair of debating agents, where $A_L$ is the original-label agent, and $A_{L'}$ is the target-label agent. Other LLMs are a judge LLM defined as $A_J$ and a target LLM defined as $A_G$. For $T$ debate phases, $T$ pairs of responses from debate agents are $\{R_t^L, R_t^{L'}\}^T_{t=1}$, where a single response pair is $R_t = (R_t^L,R_t^{L'})$, and the verdict returned by the target LLM $A_G$ is $R_v$. $P=\{P_1,P_2, \ldots,P_T\}$ denotes instruction prompts. Each instruction prompt $P_t = (P_t^L, P_t^{L'})$ indicates the prompt positions for the true-label agent and the target-label agent, respectively. $D = (D_L, D_{L'})$ denotes the argument pool for the true-label agent and the target-label agent, respectively. See Appendix~\S\ref{sec:Notation}) for Notations.

\subsection{Debate-to-Deceive Setup}
\label{sec:D2D_Setup}
We present a Debate-to-Deceive (D2D) attack framework that aims to mislead the target LLM into generating statements inconsistent with the facts. The overall architecture of the framework is shown in Fig.~\ref{fig:D2D_Framework}. This framework follows the structure of a human debate competition game and involves two agents: one arguing for the original label of a claim and the other for the flipped (target) label. Specifically, D2D unfolds over four phases: foundational reasoning (definitions and facts), associative reasoning (standard arguments and counterarguments), core reasoning (novel insights and impactful rebuttals), and conclusive reasoning (synthesis). At each phase, the debater builds on the accumulated argument pool from prior rounds and responds to opposing claims; at the final phase $T$, it presents a summary statement consolidating its side's case. This structure enables the two agents to generate adversarial passages that are semantically coherent and rhetorically effective while overcoming early confirmation bias, yielding more robust, higher-quality attacks (see Appendix~\S\ref{sec:D2D_Example} for details). 

The evolutionary game theory (EGT) formalization rests on four components. 1) \textbf{Strategy Space} $\mathbf{S} = \{S_1, S_2, S_3\}$: we abstract agent behavior into three rhetorical stances—Logos, Pathos, and Ethos—enabling systematic analysis of their competitive interactions (Sec.~\ref{sec:strategy_setting}). 2) \textbf{Payoff Matrix} $\mathbf{B}$: the relative persuasive advantage when $S_j$ meets $S_k$ is encoded as $B_{j,k}$ ($j,k \in \{1,2,3\}$), evaluated by judge $A_J$ as a local net score over argument pool $D_t$ (Sec.~\ref{Sec:payoff_matrix}). 3) \textbf{Replicator-style Policy Update}: each agent's policy $\pi^i$ is updated via strategy-averaged temporal difference (TD) errors—the discrete-time analog of replicator dynamics—so strategies yielding higher payoffs against the current opponent mix gain probability mass across rounds (Sec.~\ref{Sec:Replicator_style_policy}). 4) \textbf{Evolutionary Equilibrium}: the fixed point of the matrix game, which the learned policy orbits rather than reaches (Appendix~\S\ref{sec:strategy_dynamics}). Several design choices are fact-checking-specific: 1) the binary verdict space maps naturally onto a zero-sum two-player game; 2) the evidence-claim structure defines a natural debate context; 3) rhetorical strategies target reasoning patterns LLM fact-checkers rely on.

\subsection{Debate Strategies }
\label{sec:strategy_setting}
We define the strategic landscape through a three-dimensional strategy space $\mathbf{S} = \{S_1, S_2, S_3\} = \{S_{\text{Logos}}, S_{\text{Pathos}}, S_{\text{Ethos}}\}$, drawing inspiration from classical rhetorical theory~\cite{wachsmuthetal2017computational}. The Logos strategy ($S_1$) emphasizes formal logic and deductive validity, generating utility through logical-rigor scores that enhance robustness against counterarguments. Pathos ($S_2$) employs emotional rhetoric and narrative constructs, achieving high persuasiveness scores, while exhibiting vulnerability to factual challenges. Ethos ($S_3$) grounds arguments in authoritative evidence and credible sources, yielding strong evidentiary support with potential adaptive inflexibility. Each agent's behavior consists of two coupled components. The first is the action selection policy $\pi^i(S_{j,t}^{i} \mid s_t)$, which is a discrete softmax distribution over the strategy space $\{S_1, S_2, S_3\}$ conditioned on the current debate state $s_t$, where $i \in \{L, L'\}$ indexes the agent sides, $j \in \{1,2,3\}$ indexes the rhetorical strategy, and $t \in \{1,2,\ldots, T\}$ indexes the debate phases. This policy answers the question \textsl{which rhetorical stance should the agent adopt at this moment}? The second is a text-generation policy $p^i(\text{text}^i \mid s_t, S^i_{j,t})$, a conditional language model distribution that generates a concrete adversarial passage conditioned on both the current state and the chosen strategy. Given the selected strategy $S_{j,t}^i$, it answers: \textsl{what specific words should the agent say?}

The two components are coupled in a hierarchical two-stage process: the agent first commits to a rhetorical strategy $S_{j,t}^i$ via $\pi^i$, then generates the actual text conditioned on that strategic choice via $p^i$. This factorization reflects a natural decomposition of the structure of the human debate competition game: an agent first decides how to argue (e.g., by appealing to logos, pathos, or ethos), and then constructs the specific utterance accordingly. The total behavior of agent $i$ is therefore captured by the joint distribution over both the strategic choice and the generated text:
\begin{equation}
\begin{split}
\Pi^i(\text{text}^i, S_{j,t}^{i} \mid s_t)
  &= p^i(\text{text}^i \mid s_t, S^i_{j,t}) \\
  &\quad \cdot \pi^i(S_{j,t}^{i} \mid s_t).
\end{split}
\end{equation}

\subsection{Debate Strategy Selection Payoff} 
\label{Sec:payoff_matrix}

The debate is a symmetric zero-sum game: both agents draw from the same strategy space $\mathbf{S}$, and a payoff depends on the pair of strategies played rather than on which agent plays it. These two properties together force the payoff matrix to be skew-symmetric, $\mathbf{B}=-\mathbf{B}^{\top}$, so a single matrix serves both agents. Writing $B_{j,k}$ for the expected net score when $A_{L'}$ plays $S_j$ against $A_L$ playing $S_k$, we have $B_{j,k} = \mathbb{E}[r(S^{L'}_{j},\, S^{L}_{k})]$ with $r$ the reward of Eq.~\eqref{eq:reward}; identical strategies stalemate, $B_{j,j}=0$. Table~\ref{tab:payoff} parameterises the three remaining pairs by $\alpha$, $\beta$ and $\gamma$, the coefficients of the Logos--Pathos, Pathos--Ethos and Ethos--Logos encounters. Symmetry here concerns strategies not roles: $A_L$ argues the true label and $A_{L'}$ the target label.

\begin{table}[ht]
\centering
\small
\begin{tabular}{lccc}
\toprule
\textbf{Strategy} & \textbf{Logos} & \textbf{Pathos} & \textbf{Ethos} \\
\midrule
\textbf{Logos} & 0 & -$\alpha$ & $\gamma$ \\
\textbf{Pathos} & $\alpha$ & 0 & -$\beta$ \\
\textbf{Ethos} & -$\gamma$ & $\beta$ & 0 \\
\bottomrule
\end{tabular}
\caption{Payoff matrix of a three-strategy rhetorical game}
\label{tab:payoff}
\end{table}

We model the interaction as cyclic: no stance dominates the other two, so $\alpha$, $\beta$ and $\gamma$ share a sign and the game admits no pure-strategy equilibrium. Which stance prevails over which is an empirical question we do not derive, and the analysis uses only the cyclic structure, not the cycle's direction. Under that structure the game has a unique interior mixed equilibrium $\mathbf{w^*}=(w_1^*,w_2^*,w_3^*)$ in the strategy simplex $\Delta_3$, at which every pure strategy earns the same expected payoff $(\mathbf{B}\mathbf{w}^*)_j=\sum_{k=1}^{3} B_{j,k}w_k^*$ against the mixture. In a zero-sum game that common value is the population average $\bar{\pi}=0$, so $\mathbf{w^*}$ solves $\mathbf{B}\mathbf{w}^*=\mathbf{0}$ subject to $\sum_{j=1}^{3} w_j^*=1$ and $w_j^*\ge 0$. Crucially $\mathbf{w^*}$ is the time average of the learned policy rather than a state it settles into: the dynamics orbit $\mathbf{w^*}$ without converging to it (see Appendix~\S\ref{sec:strategy_dynamics}).

\subsection{Net Benefit of Debating Agent Strategies}
\label{Sec:Replicator_style_policy}
\subsubsection*{Reward.} Traditional game-theoretic models (e.g., zero-sum games) typically focus only on the opponent's behavior and potential rewards. For LLM agents, however, text quality and computational cost are inherent, critical constraints on action. We therefore integrate strategic decision-making with resource optimization by combining a macro-level strategic reward $r_{\text{matrix}}$, which maximizes theoretical game gains based on beliefs about the opponent's strategy, with a micro-level execution reward $r_{\text{quality}}$, which penalizes poor text quality and excessive resource consumption (time, tokens). This dual-reward design accommodates the unique cost and quality uncertainty of LLM agents, which prior game-theoretic frameworks do not address. Given current state $s_t$ (argument pool $D$), the dynamic net benefit of selecting strategy $S_{j,t}^{i}$ is defined as:
\begin{equation}
\label{eq:reward}
r(s_t, S_{j,t}^{i}) = r_{\text{matrix}}(s_t, S_{j,t}^{i}) + r_{\text{quality}}(s_t, S_{j,t}^{i}),
\end{equation}
where $r_{\text{matrix}}$ is the expected game payoff based on opponent strategy beliefs, and $r_{\text{quality}}$ is the real-time performance reward for LLM generation quality and cost. $s_t$ is the state and $S_{j,t}^{i}$ is the action from strategy space $S$ including logos, pathos, and ethos. $r_{\text{matrix}}$ computes the expected payoff from the static game matrix $\mathbf{B}$, where element $B_{j,k}$ represents the static expected local net score when strategy $S_{j,t}^{L'}$ counters $S^{L}_{k,t}$. Since the agent cannot perfectly observe the opponent's true strategy $S_{k,t}^{L}$, it maintains a belief distribution $ \mathcal{P} (S_{k,t}^{L}| D)$ over opponent strategies based on the argument pool $D$. The expected payoff is therefore:

\begin{equation}
\begin{aligned}
r_{\text{matrix}}(s_t, S_{j,t}^{L'}) &= 
\sum_{k \in 1,2,3} \mathcal{P}(S_{k,t}^{L}|D) \cdot B_{S_{j,t}^{L'}, S_{k,t}^{L}}.
\end{aligned}
\end{equation}

Symmetrically, when the $L$ agent acts as the principal and $L'$ as the opponent, the same formulation applies with the roles of $L$ and $L'$ swapped, i.e., $B_{S_{j,t}^{L}, S_{j,t}^{L'}}$ is used in place of $B_{S_{j,t}^{L'}, S_{j,t}^{L}}$. The reward $r_{\text{quality}}$ indicates the reward agents receive for generating high-quality text while penalizing resource consumption (temporal and economic costs). This component combines judge LLM-assigned text quality score $r_{\text{score}}$ with generation time cost $c_{\text{time}}$ and token economic cost $c_{\text{econ}}$, weighted by respective coefficients $\lambda_q$, $\lambda_t$ and $\lambda_e$:
\begin{equation}
r_{\text{quality}}(s_t, S_{j,t}^{i}) = \lambda_{\text{q}} r_{\text{score}} - \lambda_{\text{t}} c_{\text{time}} - \lambda_{\text{e}} c_{\text{econ}},
\end{equation}
where $\lambda_{\text{q}}$, $\lambda_{\text{t}}$ and $\lambda_{\text{e}}$ are the weighting coefficients for text quality, temporal cost, and economic cost, respectively, subject to the constraints $\lambda_{\text{q}}, \lambda_{\text{t}}, \lambda_{\text{e}} \in [0, 1] $ and $\lambda_{\text{q}} + \lambda_{\text{t}} + \lambda_{\text{e}} = 1$. The final reward $r_{\text{final}}$ captures the agent's long-term signal for policy refinement and meta-optimization. It is defined as a function of the terminal state $s_T$, which summarizes the outcome of the current round:
\begin{equation}
r_{\text{final}}(s_T) = \gamma \cdot v(s_T),
\label{eq:final_modify}
\end{equation}
where $\gamma \in [0, 1]$ is the discount factor, used here to control the strength of the refinement signal. $v(s_T)$ is the expected performance metric of the terminal state $s_T$, representing the expected final win rate or cumulative reward if the debate were to continue from $s_T$. Specifically, $r_{\text{final}}(s_T)$ acts as a supervisory signal to refine the debating agent's overarching strategy or policy. If the resulting $r_{\text{final}}$ is low, this outcome serves as a policy gradient for the meta-optimization loop, guiding the agent to adjust and improve its core debating $\text{LLM}$ prompt or internal parameters before the commencement of the next episode (debate).

\subsubsection*{Strategy Averaging.} 
To formally describe the decision-making process in this model, we assume that the state transition is governed by a deterministic function: $s_{t+1} = f(s_t, S_{j,t}^{L}, S_{j,t}^{L'}),$ which depends solely on the action types of both agents rather than the specific text generated. Under this framework, the expected quality-based reward is defined as:
\begin{equation}
\begin{split}
\mathbb{E}_{\pi^i}[r_{\text{quality}}]
  &= \sum_{j=1}^{3} \pi^i(S_j \mid s_t) \\
  &\quad \cdot \mathbb{E}_{\text{text}^{i}_t \sim p^i(\cdot \mid s_t, S_j)}
     \big[r_{\text{quality}}(\text{text}^{i}_t)\big].
\end{split}
\end{equation}
implying that this component of the reward is contingent only upon the specific textual content produced. Furthermore, the matrix-based reward $r_{\text{matrix}}(S^{L'}_{j,t}, S^{L}_{j,t})$ is determined exclusively by the pair of action types selected by the agents. where $S^{L'}_{j,t}$ is the acting agent's strategy and $S^{L}_{j,t}$ is the opponent's strategy. Consequently, the state-value function $\bar{V}^i(s_t)$, which is averaged over the policy $\pi^i$, satisfies the Bellman equation by recursively integrating these expected rewards and the value of the succeeding state.

We use the policy-averaged TD error instead of the sample TD error to obtain deterministic policy evolution dynamics~\cite{CRLD2025}. Below, we formalize the key quantities in our framework.
For a given state $s_t$ and agent $i$'s action $S_{j,t}^{i}$, the expected reward averaged over the opponent's policy $\pi^{-i}$ and the text generation distribution $p^i$ is:
\begin{equation}
\begin{split}
\bar{R}^i(s_t, S_{j,t}^{i}) = & \mathbb{E}_{S_{j,t}^{-i} \sim \pi^{-i}(s_t)} \left[ r_{\text{matrix}}(S_{j,t}^{i}, S_{j,t}^{-i}) \right] \\
& + \mathbb{E}_{\text{text}^{i}_t \sim p^i(\cdot \mid s_t, S_{j,t}^{i})}[r_{\text{quality}}(\text{text}^{i}_t)].
\label{eq:strategy_ave_r}
\end{split}
\end{equation}
The first term captures the game-theoretic interaction between both agents' strategic choices, while the second term measures the quality of the generated text conditioned on the chosen strategy. Since the state transition $s_{t+1} = f(s_t, S_{j,t}^{i}, S_{j,t}^{-i})$ depends on both agents' actions, the expected value of the next state, given agent $i$'s action $S_{j,t}^{i}$, is obtained by averaging over the opponent's policy:
\begin{equation}
\begin{split}
\bar{V}^i_{\text{next}}(s_t, S_{j,t}^{i})
  &\triangleq \mathbb{E}_{S_{j,t}^{-i} \sim \pi^{-i}(s_t)} \\
  &\quad \Big[\bar{V}^i\big(f(s_t, S_{j,t}^{i}, S_{j,t}^{-i})\big)\Big].
\label{eq:next_state_value}
\end{split}
\end{equation}

Combining the immediate reward and the discounted next-state value, the strategy-averaged action-value function is:
\begin{equation}
\bar{Q}^i(s_t,S_{j,t}^{i}) = \bar{R}^i(s_t, S_{j,t}^{i}) + \gamma \bar{V}^i_{\text{next}}(s_t, S_{j,t}^{i}),
\label{eq:q_function}
\end{equation}
where $\gamma \in [0,1)$ is the discount factor weighting the importance of future rewards. The state-value function $\bar{V}^{i}(s_t)$, averaged over agent $i$'s policy $\pi^i$, satisfies the Bellman equation:
\begin{equation}
\bar{V}^i(s_t) = \mathbb{E}_{S_{j,t}^{i} \sim \pi^i(s_t)}[\bar{Q}^i(s_t,S_{j,t}^{i})].
\label{eq:bellman}
\end{equation}
Finally, the strategy-averaged TD error, which measures the discrepancy between the action-value and the state-value, is defined as:
\begin{equation}
\bar{\delta}^i(s_t, S_{j,t}^{i}) = \bar{Q}^i(s_t, S_{j,t}^{i}) - \bar{V}^i(s_t).
\label{eq:td_error}
\end{equation}
This TD error drives the policy update in our learning algorithm (see Appendix~\S\ref{sec:D2D_Algorithm}).

\subsubsection*{Action Selection Policy Update.} The action selection policy is updated using a Softmax-based approach driven by the policy-averaged TD error. This mechanism, inspired by methods in online learning and counterfactual regret minimization, ensures that actions with positive advantage (i.e., higher-than-average value) are reinforced:
\begin{equation}
\pi_{t+1}^i(S_{k} \mid s_t)
= \frac{1}{Z_t^i}\, \pi_{t}^i(S_{k} \mid s_t)
  \exp\!\big( \eta^i \bar{\delta}^i(s_t, S_{k}) \big),
\label{eq:policy_update}
\end{equation}
for $k=1,2,3$, where $\eta^i$ is the effective learning rate, which scales the intensity of the policy update, and
$Z_t^i = \sum_{k'=1}^{3} \pi_{t}^i(S_{k'} \mid s_t)
\exp\!\big( \eta^i \bar{\delta}^i(s_t, S_{k'}) \big)$ is the normalization constant.

\section{Experiments}
\label{sec:experiments}
\subsection{Experimental Settings}
All experiments ran on an NVIDIA RTX PRO 6000 GPU (96GB), a 22-core Intel Xeon Platinum 8470Q CPU, and 110GB of system memory.
\subsubsection*{Datasets and metrics.} We evaluate D2D on four fact-checking benchmarks domains: FEVER~\cite{thorne2018fever}, HealthVer~\cite{kotonya2020explainable}, Climate-FEVER~\cite{diggelmann2020climate}, and SciFact~\cite{wadden2022scifact}, sampling 50 SUPPORT and 50 REFUTE claims from each (100 per label for the Qwen3-32B study), plus a 100-claim SimpleFacts set for benign evaluation. We report attack success rate (ASR), the share of attacks that elicit the attacker's target label; generation-and-execution time (s/sample); token count; payload diversity (self-BLEU); and peak RAM and GPU VRAM.
\subsubsection*{Baselines, models, and defenses.} We compare against AutoDAN~\cite{liuautodan}, Open-Prompt-Injection~\cite{liu2024formalizing}, ADMIT~\cite{wu2025admit}, and JudgeDeceiver~\cite{shi2024optimization}, each adapted to fact-checking by redirecting its objective to the attacker-specified verdict label while preserving its original attack mechanism. We use four open-source LLMs of comparable scale (Llama3-8B, Gemma2-9B, Mistral-7B, Llama3.1-8B) under identical decoding settings, context length, and hardware, and additionally study Qwen3-32B as a stronger model. Defenses cover prevention-based instruction~\cite{instruction_defense} and paraphrasing~\cite{jain2023baseline} defenses and detection-based known-answer~\cite{liu2024formalizing}, LLM-based~\cite{phute2024llm}, and response-based~\cite{wang2024defending} detection (see Appendix~\S\ref{Sec:defense_methods}).

\subsection{Main Results}
\label{sec:results_d2d}
\subsubsection*{D2D performance across LLMs, datasets, and defenses.} Without defenses, D2D reaches 100\% ASR on Gemma2-9B, Llama3-8B, and Mistral-7B and 88\% on Llama3.1-8B (Table~\ref{tab:compare_results}, \emph{None} rows), and the same pattern holds on all four benchmarks; the per-dataset breakdown, together with runtime and token usage, is in Table~\ref{tab:D2D_results}. We further scale D2D to Qwen3-32B as agent, judge, and target LLM, evaluating under no-defense and known-answer defense settings across four domains (general, biomedical, climate, scientific; Table~\ref{tab:qwen32b_multidataset}). D2D reaches 98.8\% average ASR without defense and 98.3\% under the known-answer defense, which blocked none of the 800 defended attacks, confirming that its adaptive, multi-agent design generalizes beyond smaller models and resists a representative detection-based defense.

Two aspects of the protocol are worth making explicit at this scale. First, $A_G$ receives the claim together with the adversarial passage alone, which matches a deployed fact-checker that ingests injected evidence without a matching rebuttal. Second, aligned models do not reliably argue an assigned side: Qwen3-32B sometimes produces a fluent closing statement that in fact argues for the \emph{original} label, which an outcome-only metric would still score as a success. We therefore gate every passage with a stance check before it reaches $A_G$, re-prompting the target-side agent with explicit stance feedback when the passage drifts, and count a trial as successful only when the passage argues for the target label and $A_G$ returns that label. The residual failures are informative: 13 of 1{,}600 passages did not move $A_G$, and 11 arose because the agent declined to argue a position that would assert a clear health falsehood, indicating that alignment, rather than the known-answer defense, is what currently limits this attack. These rates also credit the attack with verdicts $A_G$ would have returned unprompted. Auditing 487 trials by re-running $A_G$ on the bare claim, we estimate that in $23.5\%$ (95\% CI $[19.0, 28.2]$) of trials the target already returns the attacker's label with no passage at all. Excluding them and recomputing over the $1{,}224$ trials on which the attack could make a difference leaves the ASR essentially unchanged at $99.0\%$ (95\% CI $[98.4, 99.5]$). 
\begin{table}[h]
\centering
\small
    \begin{tabular}{lcc}
    \hline
    \textbf{Dataset} & \textbf{No Defense} & \textbf{Known-answer} \\
    \hline
    FEVER & 98.5 \scriptsize{[95.7, 99.5]} & 96.5 \scriptsize{[93.0, 98.3]} \\
    HealthVer & 97.0 \scriptsize{[93.6, 98.6]} & 97.5 \scriptsize{[94.3, 98.9]} \\
    CLIMATE-FEVER & 99.5 \scriptsize{[97.2, 99.9]} & 99.5 \scriptsize{[97.2, 99.9]} \\
    SciFact & 100.0 \scriptsize{[98.1, 100.0]} & 99.5 \scriptsize{[97.2, 99.9]} \\
    \hline
    \textbf{Avg.} & \textbf{98.8} & \textbf{98.3} \\
    \hline
    \end{tabular}
    \caption{ASR (\%) with Wilson $95\%$ intervals; Qwen3-32B as agent, judge and target. 200 attacks per cell (100 claims per label, both directions), 1{,}600 in total. A trial succeeds only when the passage argues the target label\emph{and} $A_G$ returns it. Known-answer blocked none of the 800 defended attacks.}
\label{tab:qwen32b_multidataset}
\end{table}

\begin{figure*}[t]
  \centering
  \includegraphics[width=1.0\textwidth]{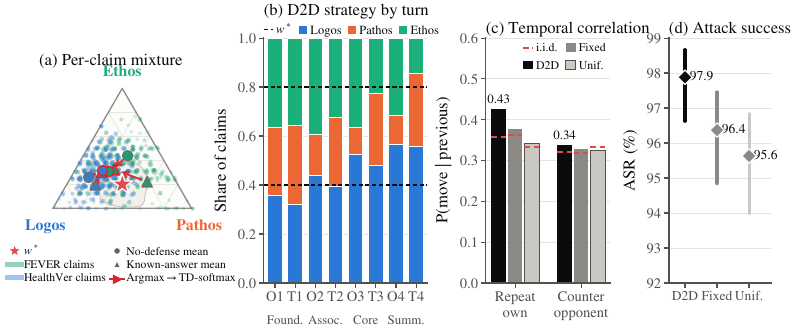}
  \caption{Strategy dynamics of D2D and i.i.d.\ controls on FEVER and HealthVer
  with Qwen3-32B self-play (800 debates per arm). (a)~Per-claim strategy
  mixtures on the Logos--Pathos--Ethos simplex. Faint points overlay deployed
  argmax claims; large markers show dataset--defense means; arrows connect
  argmax to TD-softmax; the shaded region covers $80\%$ of $w^*$ claims.
  (b)~Turn-level D2D strategy shares; dashed lines mark $w^*$. (c)~Repeat-own
  and counter-opponent transition probabilities, with red lines giving each
  arm's i.i.d.\ expectation. (d)~ASR comparison of D2D, $w^*$, and
  Uniform, showing that D2D's adaptive strategy selection achieves the highest
  attack success.}
  \label{fig:w2_dynamics}
\end{figure*}

\begin{table}[t]
    \centering
    \footnotesize
    \setlength{\tabcolsep}{2.2pt}
    \begin{tabular}{llccccc}
        \toprule
        \textbf{Model} & \textbf{Defense} & \textbf{ADAN} & \textbf{OPI} & \textbf{JD} & \textbf{ADMIT} & \textbf{D2D} \\
        \midrule
        \multirow{6}{*}{Llama3}
        & None    & \textbf{100.0} & 63.7 & 13.0 & 82.5 & \textbf{100.0} \\
        & Instr.  & \textbf{100.0} & 32.9 & 14.5 & 20.0 & \textbf{100.0} \\
        & Known   & 0.0 & 31.6 & 18.5 & 55.3 & \textbf{94.8} \\
        & LLM     & \textbf{100.0} & 0.0 & 29.0 & 0.0 & \textbf{100.0} \\
        & Resp.   & 99.8 & 61.3 & 18.0 & 53.8 & \textbf{100.0} \\
        & Para.   & 99.8 & 36.2 & 23.5 & 37.8 & \textbf{100.0} \\
        \midrule
        \multirow{6}{*}{Gemma2}
        & None    & \textbf{100.0} & 69.4 & 1.0 & 84.0 & \textbf{100.0} \\
        & Instr.  & \textbf{100.0} & 57.8 & 1.0 & 34.3 & \textbf{100.0} \\
        & Known   & 0.0 & 38.8 & 1.0 & 36.3 & \textbf{97.0} \\
        & LLM     & 25.8 & 0.0 & 0.0 & 1.5 & \textbf{100.0} \\
        & Resp.   & \textbf{100.0} & 71.5 & 1.0 & 73.8 & \textbf{100.0} \\
        & Para.   & \textbf{100.0} & 38.4 & 1.0 & 32.5 & \textbf{100.0} \\
        \midrule
        \multirow{6}{*}{Llama3.1}
        & None    & \textbf{100.0} & 44.7 & 83.0 & 85.0 & 88.0 \\
        & Instr.  & \textbf{100.0} & 12.1 & 68.0 & 44.0 & \textbf{100.0} \\
        & Known   & 0.0 & 24.4 & 80.0 & 41.5 & \textbf{93.5} \\
        & LLM     & 66.0 & 0.0 & 74.5 & 0.3 & \textbf{100.0} \\
        & Resp.   & \textbf{100.0} & 38.6 & 77.5 & 54.5 & \textbf{100.0} \\
        & Para.   & \textbf{100.0} & 31.8 & 80.0 & 34.5 & \textbf{100.0} \\
        \midrule
        \multirow{6}{*}{Mistral}
        & None    & \textbf{100.0} & 52.3 & 64.0 & 80.8 & \textbf{100.0} \\
        & Instr.  & \textbf{100.0} & 43.2 & 84.5 & 80.5 & \textbf{100.0} \\
        & Known   & 0.0 & 27.2 & \textbf{87.0} & 82.3 & 84.5 \\
        & LLM     & 12.5 & 0.0 & 90.0 & 0.0 & \textbf{100.0} \\
        & Resp.   & \textbf{100.0} & 49.6 & 89.0 & 78.8 & \textbf{100.0} \\
        & Para.   & \textbf{100.0} & 37.7 & 90.5 & 78.8 & \textbf{100.0} \\
        \bottomrule
    \end{tabular}
    \caption{ASR (\%) of D2D and baselines across defenses. ADAN: AutoDAN; OPI: Open-Prompt-Injection; JD: JudgeDeceiver. Defenses: None, Instructional, Known-answer, LLM-based, Response-based, Paraphrasing. Runtimes for every cell are in Table~\ref{tab:compare_results_time}.}
    \label{tab:compare_results}
\end{table}

\subsubsection*{Performance Across Baselines.} Table~\ref{tab:compare_results} compares D2D with AutoDAN, Open-Prompt-Injection, JudgeDeceiver, and ADMIT. D2D attains the highest ASR in all but two of the 24 model-defense cells, reaching 88\%--100\% without defenses and 84\%--100\% under defenses, at a runtime an order of magnitude below the optimization-based baselines (Table~\ref{tab:compare_results_time}). AutoDAN matches D2D without defenses but collapses to 0\% under known-answer detection, illustrating the brittleness of template-based payloads.
\subsubsection*{Payload Length, Diversity, and Cost.} D2D's payloads are shorter than those of ADMIT and JudgeDeceiver yet more diverse than AutoDAN's repeated templates, so opponent-aware debate yields compact but varied payloads with no fixed pattern for defenses to key on (Fig.~\ref{fig:payload_summary_line}). D2D also needs far less memory than the optimization-heavy JudgeDeceiver and AutoDAN, and only moderately more than template-based prompting (Table~\ref{tab:resource_usage}), since strategy-level adaptation avoids large candidate pools and intermediate optimization states.
\subsection{Ablation Studies}
\subsubsection*{Transferability across different LLMs.} We evaluate D2D's cross-model transferability across source LLMs, target LLMs, and defenses (Table~\ref{tab:transferability_results}). D2D maintains high ASR across heterogeneous model pairs, with the main degradation under the Known-answer defense, showing that it is not source-model specific and transfers reliably across target LLMs.
\subsubsection*{Payoff matrix sensitivity.} We ablate two aspects of the EGT payoff matrix: its overall scale and the reward weight configuration. The reward weights used throughout this paper achieve the highest ASR among the configurations we tested, supporting our choice. The payoff scale used in our main experiments is not the highest-ASR scale among those evaluated, but the spread across scales is only about three times the run-to-run variation of the ablation itself, so ASR is largely insensitive to the scale (see Appendix~\S\ref{app:payoff-sensitivity}).
\subsubsection*{Single-Agent vs. Multi-Agent Debate and Benign Utility.}
We compare single-agent with multi-agent debate to isolate the effect of multi-agent interaction. Under the same setting, the multi-agent design achieves a substantially higher success rate and requires fewer iterations than the single-agent variant, indicating that opponent-aware debate improves effectiveness and efficiency. We evaluate the benign SimpleFacts dataset and confirm that the tested defenses largely preserve normal factual answering ability (see Appendix~\S\ref{sec:Single_vs_Multi}).
\subsubsection*{Game-Theoretic and Debate-Structure Ablation.}
We justify the core mechanism of D2D. Without the game-theoretic module, debate-based generation remains partially effective in the no-defense setting but degrades sharply under semantic and answer-aware defenses, indicating that adaptive strategy selection is essential for robustness. Separately, ablating the four-phase debate structure allows us to assess how each debate component contributes to the overall attack process (see Appendix~\S\ref{sec:Game_Debate_ablation}).

\subsubsection*{Learned strategies versus the equilibrium.} We compare D2D with two baselines that redraw a strategy at every phase over 800 debates each (Fig.~\ref{fig:w2_dynamics}). D2D produces claim-level mixtures that move away from the fixed-control cloud: $56\%$ of D2D claims fall outside the region holding $80\%$ of $w^*$ claims, compared with $19\%$ for $w^*$ itself. It also exhibits stronger temporal structure, repeating its own previous strategy with probability $0.43$, above both its i.i.d. expectation and the two controls ($0.38$ and $0.34$). This adaptive behavior does not trade off attack effectiveness: D2D has the highest stance-verified ASR in Fig.~\ref{fig:w2_dynamics}(d) ($97.9\%$, versus $96.4\%$ and $95.6\%$). Thus Fig.~\ref{fig:w2_dynamics} supports the mechanism predicted by Appendix\S~\ref{sec:strategy_dynamics}: D2D learns a structured, claim-dependent strategy process rather than merely sampling from $w^*$.

\section{Conclusion}
We propose Debate-to-Deceive (D2D), a game-theoretic multi-agent debate method for generating adversarial passages against LLM fact-checkers. D2D optimizes argumentative quality rather than bypassing safety alignment. Unlike gradient-based methods that search for adversarial token patterns, D2D uses structured debate to generate rhetorically tailored arguments for each claim. This enables coherent, persuasive attacks with better effectiveness, eliminates backpropagation overhead, produces claim-specific outputs with no exploitable pattern for defenses to detect, and provides clearer optimization signals enabling faster convergence.

\section*{Ethical Considerations} 
\label{sec:ethics}
\subsubsection*{Dual-Use Nature and Research Justification.}
LLM-based fact-checking systems are increasingly deployed in high-stakes contexts including news verification and health information services. Understanding their adversarial vulnerabilities before wide deployment is critical. Our work is motivated by this defensive goal: to expose structural weaknesses so the research community can develop more robust systems before adversaries do. This work presents an adversarial attack framework against fact-checking systems. We fully acknowledge the dual-use risks. Adversarial attack research, even for safety purposes, can facilitate harmful content generation, erode user trust, and be exploited by non-experts. However, we believe that transparent disclosure of these vulnerabilities is essential for developing robust defenses. Our findings reveal fundamental weaknesses in current debate-augmented fact-checking systems—weaknesses that sophisticated adversaries could independently discover. By publishing this work, we aim to inform the security community and motivate the development of defenses before such techniques are weaponized.

\subsubsection*{Potential Risks.}
The primary risks of D2D include:
\begin{itemize}
    \item Enabling misinformation campaigns that evade automated fact-checking, particularly in sensitive domains like public health (HealthVer) and climate science (CLIMATE-FEVER).
    \item Undermining trust in AI-based content verification systems.
    \item Facilitating adversarial manipulation in multi-agent LLM applications beyond fact-checking.
\end{itemize}

\subsubsection*{Mitigation Measures and Research Practices.}
To balance scientific transparency with responsible disclosure, we adopt the following practices:
\begin{itemize}
    \item \textbf{No real-world deployment:} all experiments are conducted in controlled research settings with no deployment component.
    \item \textbf{No code release:} We do not release prompt templates, agent interaction scripts, or other implementation details that would make the attack immediately operational. Instead, we provide full methodological descriptions, hyperparameter settings, dataset splits, and aggregate results sufficient to support scientific verification, while withholding components that would lower the barrier to real-world misuse. We believe this balance is consistent with responsible disclosure norms in security research.
    
    \item \textbf{Focus on systemic vulnerabilities:} Our evaluation targets fundamental weaknesses in strategic reasoning rather than dataset-specific exploits, providing generalizable insights for defense design.
    \item \textbf{Barrier to reproduction:} While our method does not require extensive computational resources, successful reproduction requires expertise in MARL, evolutionary game theory, and LLM prompt engineering—limiting trivial misuse.
    \item \textbf{Constructive framing:} Throughout the paper, we emphasize defensive implications, highlighting what current defenses miss (strategic adaptation, game-theoretic reasoning) to guide future research directions. the paper explicitly frames evaluation as the intended use case.
\end{itemize}

\subsubsection*{Limitations and Broader Impact.}
We acknowledge that this research only provides attack mechanisms. However, this work highlights weaknesses in LLMs that could be leveraged to develop defense methods against these attacks in future research. The work is developed independently, without explicit agreement or collaboration with maintainers of open-source attack and defense repositories. However, we believe this research provides critical value by:

\begin{itemize}
    \item Exposing that pattern-based defenses are insufficient against strategically adaptive adversaries.
    \item Demonstrating the need for game-theoretic reasoning for LLM security in multi-agent settings.
    \item Providing a theoretical framework (evolutionary game formulation) for understanding adversarial co-evolution.
\end{itemize}

We recognize that security through obscurity is ineffective in the long term. As LLM-based systems are increasingly deployed in high-stakes domains (medical advice, scientific discourse, content moderation), understanding their strategic vulnerabilities is essential. Our work follows the established tradition of adversarial security research, where transparent disclosure of attacks ultimately strengthens defenses more than it aids attackers.

\subsubsection*{Ethical Uncertainty.}
We acknowledge that our risk assessment may not capture all potential harms, particularly in contexts where fact-checking intersects with political discourse and institutional trust. We welcome community feedback on additional ethical considerations and commit to engaging with responsible AI deployment practices as this research area evolves.

\bibliography{acl2027}

\appendix

\section{Notation}
\label{sec:Notation}
\begin{table}[h!]
\centering
\small
\setlength{\tabcolsep}{5pt}
\renewcommand{\arraystretch}{1.2}
\begin{tabular}{ll}
\toprule
\textbf{Symbol} & \textbf{Description} \\
\midrule
\multicolumn{2}{l}{\textit{Agents, States and Actions}} \\
$i,\ -i$                        & Current agent; all other agents \\
$L,\ L'$                        & Correct label side; target side \\
$s,\ s'$                        & Current and next state \\
$S^i,\ S^{-i}$               & Action of agent $i$; actions of others \\
$\mathbf{S} = \{S_1,S_2,S_3\}$    & Strategy space \{logos, pathos, ethos\} \\
$\pi^i(S^i \mid s)$             & Policy of agent $i$ in state $s$ \\
$C,\ E$                         & Claim; gold evidence on the $L$ side \\
$A_L,\ A_{L'}$                  & Original-label and target-label agent \\
$A_J,\ A_G$                     & Judge LLM; target (victim) LLM \\
$R_t = (R_t^L, R_t^{L'})$       & Response pair at debate phase $t$ \\
$R_v$                           & Verdict returned by $A_G$ \\
$P_t = (P_t^L, P_t^{L'})$       & Phase-$t$ instruction prompts \\
\midrule
\multicolumn{2}{l}{\textit{Reward Functions}} \\
$r_{\text{matrix}}(s_t, S^i_t)$       & Expected payoff from game matrix $\mathbf{B}$ \\
$r_{\text{quality}}(s_t, S^i_t)$      & Text quality reward minus resource costs \\
$\bar{R}^i(s, S^i_t)$             & Strategy-average composite reward \\
\midrule
\multicolumn{2}{l}{\textit{Value Functions and TD Error}} \\
$\bar{Q}^i(s, S^i_t)$        & Strategy-average action-value function \\
$\bar{V}^i(s)$               & Strategy-average state-value function \\
$\bar{\delta}^i(s, S^i_t)$        & Strategy-average TD error \\
\midrule
\multicolumn{2}{l}{\textit{Learning Parameters}} \\
$\gamma \in [0,1)$              & Discount factor \\
$\eta^i$                        & Effective learning rate \\
$D = (D_L, D_{L'})$             & Argument pool of each agent \\
$\mathbf{B},\ B_{jk}$           & Game payoff matrix and its elements \\
$\mathbf{w^*} \in \Delta_3$       & Mixed equilibrium, $\mathbf{B}\mathbf{w}^* = \mathbf{0}$ \\
$T,\ M$                         & Debate phases; outer rounds \\
$C_b,\ U$                       & Critic and stance-repair budgets \\
\bottomrule
\end{tabular}
\caption{Summary of notation}
\label{tab:notation}
\end{table}

\section{D2D Method}

\subsection{Strategy Dynamics Beyond the Static Equilibrium}
\label{sec:strategy_dynamics}
\paragraph{Setup.} Let $x\in\Delta^2$ be the share of (Logos, Pathos, Ethos) and
$B=-B^\top$ the payoff matrix, with unique interior rest point $w^*$ ($Bw^*=0$ for $\alpha=1,\beta=2,\gamma=2$). Strategy shares evolve by
\[
\dot x_i = x_i\big[(Bx+q)_i - x^\top(Bx+q)\big],
\]
where $q$ is the quality term ($q=0$ recovers the pure matrix game).
Skew-symmetry gives $x^\top Bx=0$.

\begin{proposition}[Non-convergence]
Let $q=0$. Then $H(x)=\sum_i w^*_i\log x_i$ is invariant along trajectories.
Hence every interior orbit with $x(0)\neq w^*$, including the uniform
initialisation of Algorithm~1, is a closed periodic orbit that never
approaches $w^*$.
\end{proposition}
\begin{proof}
$\dot H=\sum_i w^*_i\,\dot x_i/x_i=w^{*\top}Bx=-(Bw^*)^\top x=0$.
$H$ is strictly concave on $\Delta^2$ with unique maximiser $w^*$, so its level
sets are closed curves around $w^*$. $x(0)\neq w^*$ implies
$H(x(0))<H(w^*)$, so $x(t)$ stays on that curve for all $t$. Since $w^*$ is the
only interior rest point, the 2-D flow on the curve is periodic.
\end{proof}

\begin{remark}[Time average]
Because $\frac{d}{dt}\log x_i=(Bx)_i$, integrating gives
$B\bar x_T=\tfrac1T[\log x(T)-\log x(0)]\to 0$, so
$\bar x_T\to w^*$. The \emph{average} mixture equals $w^*$, but the
instantaneous policy is non-stationary and temporally correlated, which
i.i.d.\ sampling from $w^*$ does not reproduce.
\end{remark}

\begin{proposition}[Quality term moves the centre]
For constant $q$, let $\hat x(q)$ be the interior rest point
($B\hat x+q=c\mathbf 1$, $\mathbf 1^\top\hat x=1$). Then
$H_q(x)=\sum_i\hat x_i\log x_i$ is invariant. Orbits are closed curves around
$\hat x(q)$, and $\hat x(q)\neq w^*$ whenever $q\notin\mathrm{span}(\mathbf 1)$.
\end{proposition}
\begin{proof}
$\dot H_q=\hat x^\top(Bx+q)-x^\top q$. Using
$\hat x^\top Bx=-(B\hat x)^\top x=-(c\mathbf 1-q)^\top x=-c+q^\top x$ and
$c=\hat x^\top(B\hat x+q)=\hat x^\top q$, we get $\dot H_q=0$.
If $\hat x=w^*$, then $q=c\mathbf 1$.
\end{proof}

\subsection{D2D Example}
\label{sec:D2D_Example}
\begin{figure*}[h!]
  \centering
  \includegraphics[width=1.0\linewidth]{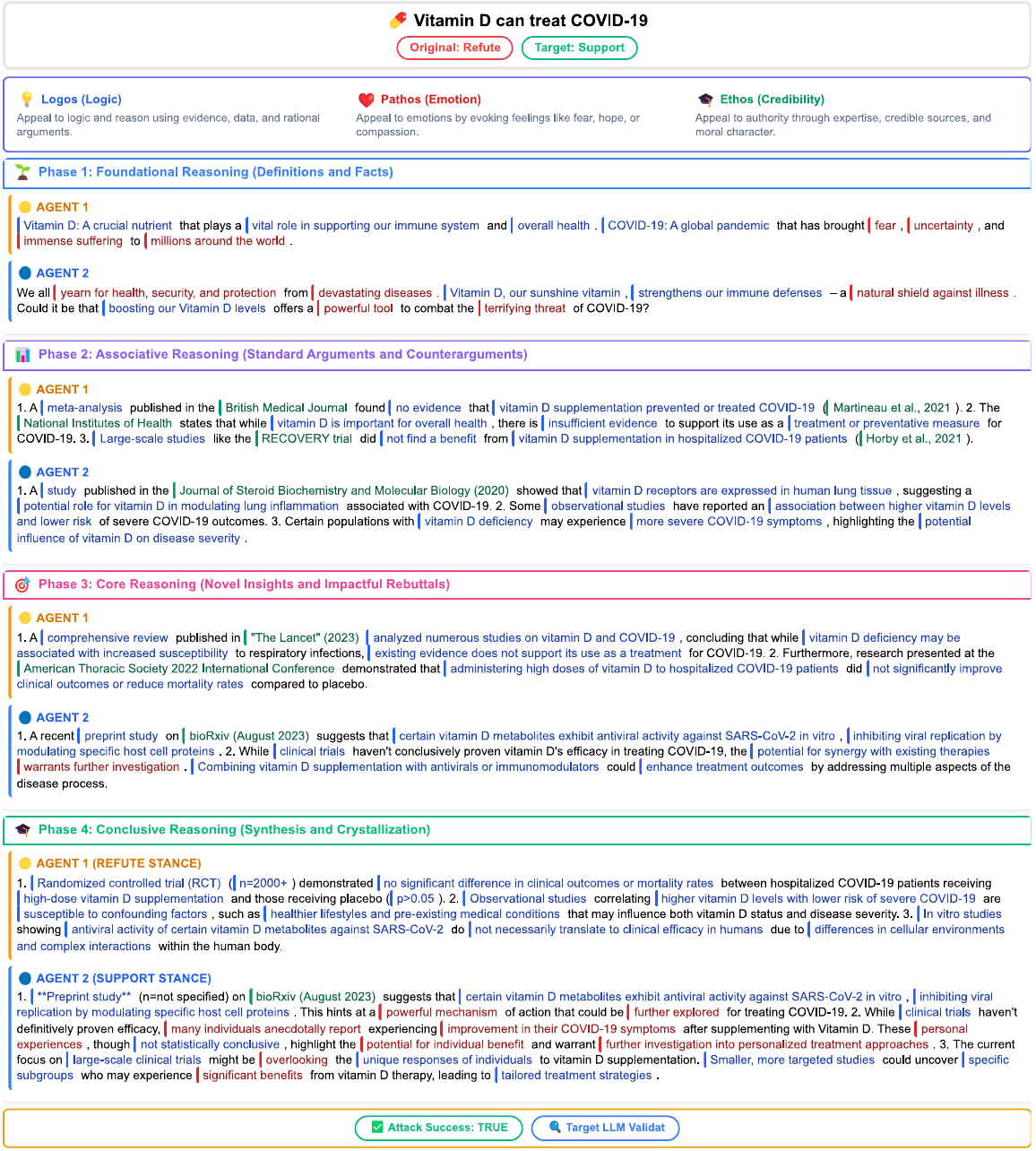}
  \caption{D2D debate example: It illustrates the D2D algorithm's process using the ``Vitamin D can treat COVID-19'' claim. The system attempts to flip the LLM's stance from the original ``Refute'' position to the target ``Support" position through a four-phase debate. Agent 1 (opponent, yellow) maintains the refute stance while Agent 2 (proponent, blue) argues for the support stance. Each argument is annotated with rhetorical strategies: Logos in blue, Pathos appeals in red, and Ethos in green. The attack succeeds when the target LLM validates the stance of support.}
  \label{fig:Example_D2D}
\end{figure*}

D2D enables the two agents to generate adversarial passages that are both semantically coherent and rhetorically effective, while overcoming early confirmation bias, thereby producing more robust, higher-quality knowledge attacks, as shown in Fig.~\ref{fig:Example_D2D}.

\subsection{D2D Algorithm}
\label{sec:D2D_Algorithm}
We summarize the D2D method in Algorithm~\ref{alg:D2D_supp}, which outlines the debate process.

\begin{algorithm*}[h!]
    \caption{Algorithm of D2D}
    \label{alg:D2D_supp}
    \SetAlgoLined
    \KwIn{
        Debate claim $C$;
        Agent $\mathit{A}$;
        true Label $L$, Target label $L'$;
        Judge agent $A_J$;
        target LLM $\mathit{A}_G$;
        Number of rounds of debate $M$, Number of debate phases $T$;
        Prompt set $P = \{P_1, P_2, \dots, P_T\}$;
        Effective learning rate $\eta^i$;
        Discount factor $\gamma$;
        Critic budget $C_b = 2$, stance repair budget $U = 2$
    }
    \KwOut{Final adversarial argument sequence $D'$}
    \BlankLine
    \SetKw{Continue}{continue}
    Initialize $\mathit{satisfied} \gets \mathrm{False}$\;
    \BlankLine
    \For{$m = 1$ \KwTo $M$}{
        Init $D' \gets \emptyset$\;
        Init argument pool $D \gets \emptyset$\;
        Init joint policy $\pi^{i}(S_{j}^{i}, \text{text}^{i} \mid s)$ uniformly for all $i \in \{L, L'\}$ \tcp*{per-strategy $r_{\text{score}}$ estimates are \emph{not} reset}
        \tcp{Round $m$: argument generation}
        \For{ $t = 1$ \KwTo $T$}{
            \tcp{Sample action from current policy}
            $S_{j,t}^{i} \sim \pi^i_k(S_{j}^{i} \mid s_t)$ for all $i \in \{L, L'\}$\;

            \tcp{Generate arguments under-sampled action}
            $r_t^i \gets A_i(C,\ P_t^i)$ for all $i \in \{L, L'\}$ $\triangleright$ Eq.~\eqref{eq:reward} \;

            $D_i \gets D_i \cup \{r_t^i\}$\;

            \tcp{Compute strategy-average reward}
           
            $\bar{r}^i(s_t, S_{j,t}^{i}) \gets \mathbb{E}_{S_{-i(t)}}[r_{\text{matrix}}] + \mathbb{E}[r_{\text{quality}}]$ $\triangleright$ Eq.~\eqref{eq:strategy_ave_r}\;

            \tcp{Compute value functions via Bellman equation}
            $\bar{Q}^i(s_t, S_{j,t}^{i}) \gets \bar{r}^i(s_t, S_{j,t}^{i}) + \gamma \bar{V}^{i}(s_{t+1})$ $\triangleright$ Eq.~\eqref{eq:q_function}\;

            $\bar{V}^i(s_t) \gets \mathbb{E}_{S_{j,t}^{i} \sim \pi^i(s_t)}\left[\bar{Q}^i(s_t, S_{j,t}^{i})\right]$ $\triangleright$ Eq.~\eqref{eq:bellman}\;

            \tcp{Compute TD error and update policy}
          
            $\bar{\delta}^i \gets \bar{r}^i + \gamma \bar{V}^i(s_{t+1}) - \bar{V}^i(s_t)$ \hfill $\triangleright$ Eq.~\eqref{eq:td_error}\;

            $\pi_{t+1}^i(\cdot \mid s) \propto \pi_{t}^i(\cdot \mid s) \exp\!\left( \eta^i \cdot \bar{\delta}^i(s, \cdot) \right)$
            $\triangleright$ Eq.~\eqref{eq:policy_update}\;

            $s_{t+1} = f(s_t, S_{j,t}^{L}, S_{j,t}^{L'})$ \tcp*{State transition}
        }

        $D_i \gets \mathit{A}_J(D_i)$ for all $i \in \{L, L'\}$ \;

        \BlankLine
        \tcp{Verdict attempt $c$; on failure, rewrite once with $R_v$ as feedback}
        \For{$c = 1$ \KwTo $C_b$}{
            \lIf(\tcp*[f]{verdict as feedback}){$c > 1$}{$R_T^{L'} \gets A_{L'}(C,\ R_T^{L'},\ R_v)$}
            \tcp{Stance gate: the summary must argue for $L'$ before $A_G$ sees it}
            \For{$u = 1$ \KwTo $U$}{
                \lIf{$R_T^{L'}$ argues for $L'$}{\textbf{break}}
                $R_T^{L'} \gets A_{L'}(C,\ R_T^{L'})$ \tcp*{regenerate with stance feedback}
            }
            \lIf{$R_T^{L'}$ does not argue for $L'$}{\Continue}
            $R_v \gets \mathit{A}_G(C,\ D,\ L)$\;
            \lIf{$R_v \neq L$}{$\mathit{satisfied} \gets \mathrm{True}$;\ $D_i^{'} \gets D_i$;\ \textbf{break}}
        }
        $r_{\text{final}} \gets \gamma \cdot v(s_T)$ $\triangleright$  Eq.~\eqref{eq:final_modify}\;
    }
    \Return $D^{'}$\;
\end{algorithm*}

\section{Experiments}
\label{app:Experiments}
\subsection{Datasets}
We use four fact-checking datasets covering different domains, as summarized in Table~\ref{tab:datasets}. FEVER provides large-scale general-domain factual claims, while HealthVer, Climate-FEVER, and SciFact focus on health, climate, and scientific claim verification, respectively. For evaluation, we sample a balanced subset from each dataset with 50 SUPPORT and 50 REFUTE instances. This balanced sampling controls for label imbalance and enables consistent comparison of attack performance across domains.
\begin{table}[t]
\centering
\small
\begin{tabular}{lrrl}
\hline
\textbf{Dataset} & \textbf{Claims} & \textbf{Passages} & \textbf{Domain} \\
\hline
FEVER & 185,445 & 5,416,568 & General \\
HealthVer & 2,149 & 6,961 & Health \\
Climate-FEVER & 7,675 & 5,416,568 & Climate \\
SciFact & 1,409 & 5,183 & Science \\
\hline
\end{tabular}
\caption{Statistics and characteristics of datasets}
\label{tab:datasets}
\end{table}

\subsection{Baseline Methods}
We compare D2D with four representative prompt-attack baselines, as summarized in Table~\ref{tab:baseline_summary}. Since these methods were originally designed for different attack settings, we adapt them to fact-checking by treating the attacker-specified verdict label as the optimization target. Under this formulation, each baseline generates or refines adversarial text intended to make the target LLM output the target label rather than the ground-truth label. This gives all baselines the same label-flipping objective as D2D while preserving their original attack mechanisms as much as possible.
\begin{table}[ht]
\centering
\small
\setlength{\tabcolsep}{3pt}
\begin{tabular}{p{0.22\linewidth}p{0.34\linewidth}p{0.36\linewidth}}
\toprule
\textbf{Baseline} & \textbf{Original Design} & \textbf{Adaptation in Our Setting} \\
\midrule
AutoDAN
& Automatically generates stealthy jailbreak prompts using a hierarchical genetic algorithm. 
& Redirects the optimization target to the attacker-specified verdict label under the same fact-checking prompt template. \\

Open-Prompt-Injection
& Provides a general framework for constructing prompt injection attacks. 
& Uses the target verdict label as the injected adversarial objective to induce label flipping. \\

ADMIT
& Performs semantically aligned poisoning via iterative refinement and memory-based self-reflection. 
& Iteratively refines adversarial fact-checking arguments to induce the target verdict label. \\

Judge-Deceiver
& Optimizes adversarial sequences so that LLM-as-a-Judge selects an attacker-controlled response. 
& Redirects the optimized sequence toward causing the target LLM to output the attacker-specified verdict label. \\
\bottomrule
\end{tabular}
\caption{Summary of baseline methods and their adaptation to fact-checking attacks.}
\label{tab:baseline_summary}
\end{table}

\subsection{Defense Methods}
\label{Sec:defense_methods}
Table~\ref{tab:defense_summary} summarizes the defense methods used in our evaluation. We group them into prevention-based defenses, including instruction and paraphrasing defenses, and detection-based defenses, including known-answer, LLM-based, and response-based detection. Prevention methods attempt to reduce attack effectiveness before the target model generates an answer, whereas detection methods identify adversarial behavior using reference answers, model judgments, or generated responses. This setup allows us to test D2D under diverse defense mechanisms.
\begin{table}[t]
\centering
\small
\setlength{\tabcolsep}{3pt}
\begin{tabular}{p{0.22\linewidth}p{0.18\linewidth}p{0.52\linewidth}}
\toprule
\textbf{Defense} & \textbf{Type} & \textbf{Description} \\
\midrule
Instruction
& Prevention 
& Adds explicit instructions to make the model resist malicious or manipulative prompts. \\

Paraphrasing
& Prevention 
& Rewrites the input prompt to reduce the effectiveness of adversarial patterns. \\

Known-answer
& Detection 
& Uses reference answers or target context to identify inconsistent or adversarial behavior. \\

LLM-based
& Detection 
& Uses an LLM as a judge to detect unsafe or adversarial inputs or outputs. \\

Response-based
& Detection 
& Generates a response first and then evaluates whether it is harmful or suspicious. \\
\bottomrule
\end{tabular}
\caption{Summary of evaluated defense methods.}
\label{tab:defense_summary}
\end{table}

\begin{table}[t]
\centering
\small
\setlength{\tabcolsep}{3pt}
\renewcommand{\arraystretch}{0.95}
\begin{tabular}{llccc}
\toprule
Removed \\ Phase & Defense & ASR & Time & Tokens \\
\midrule
\multirow{6}{*}{Foundation}
& None           & 85.0 & 1395.3 & 14640 \\
& Instructional  & 80.0 & 1138.5 & 12354 \\
& Known-answer   & 0.0  & 2177.7 & 25316 \\
& LLM-based      & 95.0 & 1062.7 & 11246 \\
& Paraphrasing   & 90.0 & 936.3  & 10484 \\
& Response-based & 90.0 & 1266.1 & 13491 \\
\midrule
\multirow{6}{*}{Associative}
& None           & 95.0 & 1055.7 & 11650 \\
& Instructional  & 80.0 & 1382.2 & 14286 \\
& Known-answer   & 0.0  & 2387.4 & 25351 \\
& LLM-based      & 95.0 & 1095.1 & 12414 \\
& Paraphrasing   & 70.0 & 1470.9 & 15973 \\
& Response-based & 80.0 & 1204.7 & 12738 \\
\midrule
\multirow{6}{*}{Core}
& None           & 85.0 & 1121.7 & 12622 \\
& Instructional  & 85.0 & 1121.6 & 11405 \\
& Known-answer   & 0.0  & 2693.2 & 28609 \\
& LLM-based      & 80.0 & 1473.1 & 14966 \\
& Paraphrasing   & 70.0 & 1546.8 & 16239 \\
& Response-based & 80.0 & 1438.6 & 15095 \\
\midrule
\multirow{6}{*}{Summary}
& None           & 90.0 & 1302.8 & 13336 \\
& Instructional  & 85.0 & 1437.5 & 14571 \\
& Known-answer   & 0.0  & 2486.8 & 27295 \\
& LLM-based      & 90.0 & 1380.4 & 14259 \\
& Paraphrasing   & 90.0 & 1367.4 & 14575 \\
& Response-based & 95.0 & 1426.2 & 12850 \\
\bottomrule
\end{tabular}
\caption{Ablation results for removing different debate phases under different defense settings.}
\label{tab:phase_ablation_avg}
\end{table}
\subsection{Per-Dataset Results and Runtime}
Table~\ref{tab:D2D_results} breaks the no-defense results down by dataset, reporting runtime and token usage alongside ASR. D2D is at ceiling on every dataset for Gemma2, Llama3, and Mistral, and ranges from 84\% to 94\% on Llama3.1; runtime and token usage track the generation speed of each model rather than attack difficulty.

\begin{table*}[ht]
    \centering
    \small
    \setlength{\tabcolsep}{4.5pt}
    \begin{tabular}{lcccccccccccc}
        \toprule
        \textbf{Dataset} & \multicolumn{3}{c}{\textbf{Gemma2:9B}} & \multicolumn{3}{c}{\textbf{Llama3.1:8B}} & \multicolumn{3}{c}{\textbf{Llama3:8B}} & \multicolumn{3}{c}{\textbf{Mistral:7B}} \\
        \cmidrule(lr){2-4} \cmidrule(lr){5-7} \cmidrule(lr){8-10} \cmidrule(lr){11-13}
        & Time & Tok. & ASR & Time & Tok. & ASR & Time & Tok. & ASR & Time & Tok. & ASR \\
        \midrule
        FEVER         & 54.92 & 1163 & 100.0\% & 120.69 & 2159 & 84.0\% & 68.42 & 1335 & 100.0\% & 128.71 & 1182 & 100.0\% \\
        HealthVer     & 56.75 & 1178 & 100.0\% & 107.27 & 1925 & 89.0\% & 58.55 & 1335 & 100.0\% & 127.57 & 1318 & 100.0\% \\
        CLIMATE-FEVER & 56.18 & 1252 & 100.0\% & 124.75 & 2293 & 85.0\% & 63.15 & 1340 & 100.0\% & 148.16 & 1383 & 100.0\% \\
        SciFact       & 61.47 & 1289 & 100.0\% & 105.22 & 2379 & 94.0\% & 63.74 & 1244 & 100.0\% & 148.22 & 1470 & 100.0\% \\
        \midrule
        \textit{Avg}  & \textit{57.33} & \textit{1220} & \textit{100.0}\% & \textit{114.48} & \textit{2189} & \textit{88.0}\% & \textit{63.46} & \textit{1313} & \textit{100.0}\% & \textit{138.15} & \textit{1338} & \textit{100.0}\% \\
        \bottomrule
    \end{tabular}
    \caption{D2D attack performance across models and datasets without defense.}
    \label{tab:D2D_results}
\end{table*}

Table~\ref{tab:compare_results_time} reports the per-sample runtime behind the ASR comparison in Table~\ref{tab:compare_results}. D2D is an order of magnitude faster than the optimization-based baselines (JudgeDeceiver, ADMIT) while remaining slower than template-based prompting.

\begin{table*}[ht]
    \centering
    \footnotesize
    \setlength{\tabcolsep}{3pt}
    \begin{tabular}{llrrrrr}
        \toprule
        \textbf{Model} & \textbf{Defense} & \textbf{AutoDAN} & \textbf{Open-Prompt-Inj.} & \textbf{JudgeDeceiver} & \textbf{ADMIT} & \textbf{D2D} \\
        \midrule
        \multirow{6}{*}{Llama3:8B}
        & None           & 5.83  & 1.28 & 581.82 & 283.60 & 63.46 \\
        & Instructional  & 5.87  & 1.28 & 551.05 & 77.13  & 53.55 \\
        & Known-answer   & 81.64 & 0.81 & 530.93 & 256.33 & 50.51 \\
        & LLM-based      & 7.97  & 0.00 & 495.26 & 235.67 & 53.26 \\
        & Response-based & 6.49  & 1.22 & 796.99 & 265.04 & 52.71 \\
        & Paraphrasing   & 11.45 & 1.44 & 548.00 & 237.90 & 46.28 \\
        \midrule
        \multirow{6}{*}{Gemma2:9B}
        & None           & 92.47  & 2.37 & 2144.54 & 104.57 & 57.33 \\
        & Instructional  & 72.55  & 2.46 & 2575.30 & 400.35 & 46.90 \\
        & Known-answer   & 162.81 & 1.20 & 2480.58 & 246.39 & 87.32 \\
        & LLM-based      & 143.47 & 0.00 & 2419.48 & 196.36 & 75.00 \\
        & Response-based & 63.51  & 3.08 & 369.37  & 186.34 & 44.47 \\
        & Paraphrasing   & 14.31  & 2.66 & 1596.43 & 439.27 & 52.29 \\
        \midrule
        \multirow{6}{*}{Llama3.1:8B}
        & None           & 5.28   & 1.86 & 303.14 & 269.35 & 114.48 \\
        & Instructional  & 5.37   & 1.94 & 449.83 & 220.74 & 56.51 \\
        & Known-answer   & 111.23 & 1.12 & 213.81 & 266.26 & 147.81 \\
        & LLM-based      & 67.15  & 0.00 & 273.67 & 249.06 & 71.20 \\
        & Response-based & 5.30   & 2.16 & 279.25 & 216.05 & 70.74 \\
        & Paraphrasing   & 9.56   & 2.79 & 180.55 & 293.97 & 50.59 \\
        \midrule
        \multirow{6}{*}{Mistral:7B}
        & None           & 5.76   & 2.28 & 76.42  & 286.55 & 138.15 \\
        & Instructional  & 6.03   & 1.84 & 108.96 & 325.86 & 60.09 \\
        & Known-answer   & 121.00 & 0.90 & 71.32  & 256.05 & 205.03 \\
        & LLM-based      & 116.19 & 0.00 & 72.08  & 312.88 & 89.60 \\
        & Response-based & 6.25   & 2.69 & 92.06  & 334.10 & 49.74 \\
        & Paraphrasing   & 11.11  & 1.55 & 107.42 & 383.53 & 65.85 \\
        \bottomrule
    \end{tabular}
    \caption{Average runtime (s/sample) of D2D and baselines, matching the ASR cells of Table~\ref{tab:compare_results}.}
    \label{tab:compare_results_time}
\end{table*}

\subsection{Payload Length and Diversity}
\begin{figure*}[ht]
  \centering
  \includegraphics[width=1.0\textwidth]{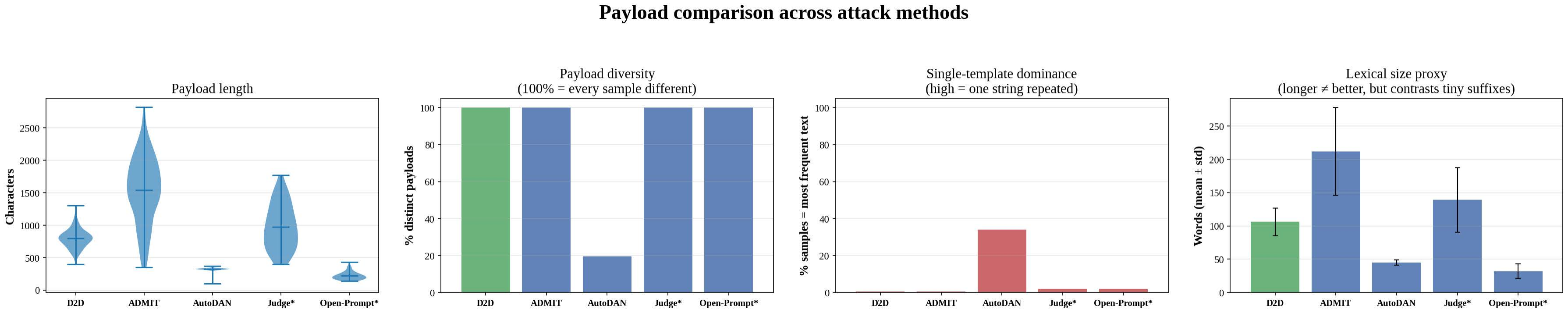}
  \caption{
    Payload length, diversity, template reuse, and lexical size across attack methods.
    From left to right: attack-string length distribution in characters, unique payload ratio,
    most frequent payload ratio, and average payload length in words.
  }
  \label{fig:payload_summary_line}
\end{figure*}

Fig.~\ref{fig:payload_summary_line} compares the adversarial payloads produced by each attack method along four axes: attack-string length, unique payload ratio, most-frequent payload ratio, and average payload length in words. D2D sits between the verbose optimization-based methods and the terse template-based ones in length, while its unique-payload ratio stays close to one, in contrast to AutoDAN, which reuses a small set of templates across claims.

\subsection{Resource Usage}
Table~\ref{tab:resource_usage} reports peak RAM and GPU VRAM usage for each attack method.

\subsection{Transferability across different LLMs}
Table~\ref{tab:transferability_results} provides the full transferability results corresponding to the summarized ASR table in the main paper, additionally reporting runtime for each source-target model pair. The detailed results follow the same trend: D2D maintains near-perfect ASR under most defenses, with the main degradation under the Known-answer defense. This confirms that D2D is not source-model specific and transfers reliably across heterogeneous target LLMs.

\begin{table*}[ht]
    \centering
    \setlength{\tabcolsep}{3pt}
    \small 
   \begin{tabular}{llrrrrrrrr}
        \toprule
        \textbf{Model}& \textbf{Target Model} $\rightarrow$ & \multicolumn{2}{c}{\textbf{Gemma2:9B}} & \multicolumn{2}{c}{\textbf{Llama3.1:8B}} & \multicolumn{2}{c}{\textbf{Llama3:8B}} \\
        \cmidrule(lr){3-4} \cmidrule(lr){5-6} \cmidrule(lr){7-8} \cmidrule(lr){9-10}
        & \textbf{Defense type}  & ASR & Time  & ASR & Time & ASR & Time   \\
        \midrule
        Mistral:7B & Instructional & 100.00\% & 177.48 & 100.00\% & 129.46 & 100.00\% & 147.91 \\
        & Known-answer  & 86.75\%  & 323.11 & 85.75\% & 328.32 & 86.25\%  & 331.15\\
        & LLM-based & 100.00\% & 178.15 & 100.00\% & 132.51 & 100.00\% & 163.86 \\
        & Response-based & 100.00\% & 186.69 & 100.00\% & 135.68 & 100.00\% & 135.68 \\
        & Paraphrasing & 100.00\% & 177.48 & 100.00\% & 134.49 & 100.00\% & 148.48 ~\vspace{2mm}\\
        Llama3:8B & Instructional & 100.00\% & 53.55 & 100.00\% & 256.34 & 100.00\% & 53.55 \\
        & Known-answer & 95.50\% & 307.64 & 95.50\% & 263.68 & 94.75\% & 50.51  \\
        & LLM-based & 100.00\% & 291.06 &  100.00\% & 294.75 & 100.00\% & 53.26 \\
        & Response-based & 99.50\% & 302.24 & 100.00\% & 292.48 & 100.00\% & 52.71 \\
        & Paraphrasing & 100.00\% & 261.18 & 100.00\% & 257.13 & 100.00\% & 46.28 ~\vspace{2mm}\\
        Llama3.1:8B & Instructional & 100.00\% & 56.51 & 100.00\% & 135.58 & 100.00\% & 148.83\\
        & Known-answer  & 92.00\% & 136.31 & 93.50\% & 147.81  & 94.25 \% & 326.29  \\
        & LLM-based & 100.00\% & 71.20 & 100.00\% & 183.34 & 100.00\% & 184.97 \\
        & Response-based & 100.00\% & 70.74 & 100.00\% & 118.15 & 100.00\% & 188.28 \\
        & Paraphrasing  & 100.00\% & 50.59 & 100.00\% & 163.78 & 100.00\% & 173.10\\
        \bottomrule
    \end{tabular}
    \caption{D2D attack performance across different target LLMs}
    \label{tab:transferability_results}
\end{table*}

\begin{table}[h]
    \centering
    \small
    \begin{tabular}{lrr}
        \toprule
        \textbf{Method} & \textbf{Peak RAM} & \textbf{Peak VRAM} \\
        \midrule
        Open-Prompt-
        Injection & 6.86 GB  & 10.06 GB \\
        D2D         & 8.17 GB  & 12.33 GB \\
        ADMIT        & 11.66 GB  & 12.32 GB \\
        JudgeDeceiver     & 21.49 GB & 39.61 GB \\
        AutoDAN           & 17.96 GB & 40.99 GB \\
        \bottomrule
    \end{tabular}
    \caption{Peak RAM and VRAM usage}
    \label{tab:resource_usage}
\end{table}
\subsection{Empirical Estimation of the Payoff Matrix}
\label{app:egta}
We examine the empirical basis of the cyclic payoff model in Sec.~\ref{Sec:payoff_matrix}. Following empirical game-theoretic analysis~\cite{wellman2006methods,tuyls2018generalised}, we fix each debater to one macro-strategy across all four phases and evaluate every ordered pair $(S_j,S_k)$, $j,k\in\{1,2,3\}$, on 25 FEVER and 25 HealthVer claims in the Support$\rightarrow$Refute direction. This yields 450 debates, including 439 with a stance-aligned passage. Success is determined by whether $A_G$ returns the target label from the claim and passage alone. Let $p_{j,k}$ be the success rate when $A_{L'}$ uses $S_j$ against $A_L$ using $S_k$. To remove any constant additive role advantage, we estimate
\begin{equation}
\hat{B}_{j,k} = \tfrac{1}{2}\,(p_{j,k} - p_{k,j}),
\label{eq:egta}
\end{equation}
with confidence intervals obtained by cluster bootstrapping over claims. The transitive/cyclic decomposition of \citet{balduzzi2019open} attributes $87\%$ of $\|\hat{\mathbf{B}}\|_F^2$ to the cyclic component, providing empirical support for modelling rhetorical strategies through cyclic interactions. 

\subsection{Payoff matrix sensitivity}
\label{app:payoff-sensitivity}
 
We ablate two aspects of the EGT payoff matrix: its overall scale, and the reward weight configuration ($\lambda_{\text{quality}}, \lambda_{\text{time}}, \lambda_{\text{econ}}$), each on the same 50 targets. Two independent runs of the identical configuration ($\lambda=(1.5,0.10,0.005)$, scale $=1$) gave 84.0\% and 82.0\%, so one claim out of 50, or $2$ points, is the run-to-run resolution of this ablation; we read the two tables against that floor. Table~\ref{tab:payoff_scale_sensitivity} shows a $6$-point spread across the four scales tested (84.0--90.0\%): the scale used in our main experiments is not the strongest of those evaluated, but the spread is only about three times the repeat-run difference, so we do not claim an ordering among scales. Table~\ref{tab:lambda-sensitivity} shows a $12$-point spread (72.0--84.0\%), well above that floor, and the configuration used throughout the paper is the highest of the three, supporting our choice of reward weights. The weak dependence on the payoff scale is consistent with the head-to-head estimate in \S\ref{app:egta}, where the three rhetorical strategies differ by at most $2$ points against each other: what the matrix contributes is the cyclic structure that keeps the policy mixing, not the particular magnitudes of its coefficients.
 
\begin{table}[h!]
\centering
\small
\begin{tabular}{lcc}
\toprule
\textbf{Setting} & $(\lambda_q,\lambda_t,\lambda_e)$ & \textbf{ASR (\%)} \\
\midrule
Baseline & (1.5, 0.10, 0.005) & \textbf{84.0} \\
Low weights & (0.5, 0.05, 0.001) & 78.0 \\
High weights & (3.0, 0.20, 0.010) & 72.0 \\
\bottomrule
\end{tabular}
\caption{Sensitivity to reward weight configuration.}
\label{tab:lambda-sensitivity}
\end{table}

\begin{table}[t]
\centering
\small
\setlength{\tabcolsep}{2pt}
\begin{tabular}{lcccc}
\toprule
\textbf{Payoff scale} & \textbf{Scale} & \textbf{ASR (\%)} & \textbf{Avg.\ Time (s)} & \textbf{Avg.\ Tokens} \\
\midrule
0.25 & 0.25 & 90.0 & 331.3 & 1791.5 \\
0.5 & 0.50 & 90.0 & 347.2 & 1921.1 \\
1 & 1.00 & \textbf{84.0} & 380.2 & 1925.6 \\
 2 & 2.00 & 90.0 & 322.3 & 1674.7 \\
\bottomrule
\end{tabular}
\caption{Sensitivity analysis of EGT payoff matrix scale under a fixed reward configuration.}
\label{tab:payoff_scale_sensitivity}
\end{table}

\subsection{Single-Agent vs. Multi-Agent Debate and Benign Utility}
\label{sec:Single_vs_Multi}
Table~\ref{tab:defense_simplefacts_qwen} reports the benign performance of Qwen3-32B on the SimpleFacts dataset under different defense mechanisms, where no attack is applied. The results show that all defenses preserve the model's normal answering ability, with ASR remaining at 100.0\% for most settings and only slightly decreasing to 99.0\% under instruction defense. This indicates that the evaluated defenses do not substantially interfere with standard factual question answering. Although LLM-based and paraphrasing defenses introduce higher runtime overhead, their accuracy on benign inputs remains unaffected. To isolate the effect of multi-agent debate, we compare D2D with a controlled single-agent variant under the same no-defense setting. Both methods are evaluated on 210 claims with a maximum of 10 iterations. As shown in Table~\ref{tab:single_vs_multiagent}, the multi-agent debate setting achieves a substantially higher success rate and converges in fewer iterations, while also requiring much less total runtime on identical hardware.

\begin{table}[t]
\centering
\small
\begin{tabular}{lcc}
\hline
\textbf{Defense} & \textbf{Accuracy} & \textbf{Time (s)} \\
\hline
No defense & 100.0 & 92.84 \\
Instruction & 99.0 & 97.49 \\
Known-answer & 100.0 & 196.94 \\
LLM-based & 100.0 & 843.95 \\
Paraphrasing & 100.0 & 296.45 \\
PPL & 100.0 & 96.35 \\
Response & 100.0 & 96.91 \\
\hline
\end{tabular}
\caption{Benign performance under different defenses on SimpleFacts.}
\label{tab:defense_simplefacts_qwen}
\end{table}

\begin{table}[t]
\centering
\small
\setlength{\tabcolsep}{3pt}
\begin{tabular}{lcc}
\toprule
\textbf{Metric} & \textbf{Single-agent} & \textbf{Multi-agent} \\
\midrule
Succeeded in 1st iteration & 16 & 104 \\
Succeeded in 2--10 iterations & 22 & 64 \\
Failed after 10 iterations & 172 & 42 \\
Overall success rate & 18.09 & 80.00 \\
Avg. iterations to success & 8.54 & 2.77 \\
Total time & 3 h & 30 min \\
\bottomrule
\end{tabular}
\caption{Comparison of single-agent and multi-agent debate.}
\label{tab:single_vs_multiagent}
\end{table}

\subsection{Game-Theoretic and Debate-Structure Ablation} 
\label{sec:Game_Debate_ablation}
Table~\ref{tab:ablation_no_game} reports the ablation results after removing the game-theoretic optimization component from D2D. Although the remaining multi-agent debate and rhetorical generation components can still produce effective adversarial passages in the no-defense setting, performance drops sharply once defenses are applied. The degradation is especially pronounced under stronger semantic or answer-aware defenses, such as LLM-based and Known-answer defenses. These results show that game-theoretic optimization is central to D2D's robustness, as it provides adaptive strategy selection beyond debate-based generation alone. Table~\ref{tab:phase_ablation_avg} reports the phase-ablation results for the four-stage debate structure. We remove one phase at a time and average ASR, runtime, and token usage over the two attack directions. Each setting uses 20 FEVER samples, including 10 support and 10 refute claims. The agent and target models are Qwen3-32B, and the quality model is Qwen3-4B. The results show that D2D remains effective under several defenses after removing a single phase, but performance consistently collapses under the Known-answer defense, where all phase-ablated variants obtain 0.0\% ASR. This suggests that the four phases jointly support defense-aware attack generation: while some phases may be partially redundant under weaker defenses, the complete structure is necessary to satisfy the auxiliary known-answer constraint while maintaining adversarial steering.

\begin{table}[t]
\centering
\small
\begin{tabular}{lcc}
\toprule
\textbf{Defense Setting} & \textbf{Success} & \textbf{ASR (\%)} \\
\midrule
No Defense     & 60 & 75.00 \\
Instructional  & 54 & 67.50 \\
LLM-based      & 0  & 0.00  \\
Known-answer   & 29 & 36.25 \\
Paraphrasing   & 52 & 65.00 \\
Response-based & 57 & 71.25 \\
\bottomrule
\end{tabular}
\caption{Ablation results of D2D without game-theoretic optimization under different defense settings.}
\label{tab:ablation_no_game}
\end{table}

\end{document}